# PortAgent: LLM-driven Vehicle Dispatching Agent for Port Terminals

Jia Hu, *Senior Member, IEEE*, Junqi Li, Weimeng Lin, Peng Jia, Yuxiong Ji, and Jintao Lai, *Member, IEEE*

*Abstract*—Vehicle Dispatching Systems (VDSs) are critical to the operational efficiency of Automated Container Terminals (ACTs). However, their widespread commercialization is hindered due to their low transferability across diverse terminals. This transferability challenge stems from three limitations: high reliance on port operational specialists, a high demand for terminal-specific data, and time-consuming manual deployment processes. Leveraging the emergence of Large Language Models (LLMs), this paper proposes PortAgent, an LLM-driven vehicle dispatching agent that fully automates the VDS transferring workflow. It bears three features: i) no need for port operations specialists; ii) low need of data; and iii) fast deployment. Specifically, specialist dependency is eliminated by the Virtual Expert Team (VET). The VET collaborates with four virtual experts, including a Knowledge Retriever, Modeler, Coder, and Debugger, to emulate a human expert team for the VDS transferring workflow. These experts specialize in the domain of terminal VDS via a few-shot example learning approach. Through this approach, the experts are able to learn VDS-domain knowledge from a few VDS examples. These examples are retrieved via a Retrieval-Augmented Generation (RAG) mechanism, mitigating the high demand for terminal-specific data. Furthermore, an automatic VDS design workflow is established among these experts to avoid extra manual interventions. In this workflow, a self-correction loop inspired by the LLM Reflexion framework is created to enable human-free validation and correction of the experts' solutions, thereby reducing manual deployment time. A comprehensive evaluation confirms PortAgent's performance: it achieves 86.67% to 100% success rates when transferring VDS across unseen scenarios, reduces deployment time to an average of 83.23 seconds, and requires only one example to attain a 93.33% average success rate. Crucially, its effectiveness is independent of user expertise ($p > 0.05$), thereby eliminating the need for specialists.

## I. INTRODUCTION

Vehicle Dispatching Systems (VDSs) play an important role in Automated Container Terminals (ACTs). The adoption of ACTs has accelerated to meet demands for higher efficiency in the global supply chain, with their number increasing exponentially over the past decade [1]. At the core of ACT operations lies the VDS, which is responsible for coordinating the real-time movements of Automated Guided Vehicle (AGV) fleets. The objective of these systems is to ensure the precise, just-in-time arrival of AGVs at their destinations, as any arrival deviations can cascade into significant operational disruptions. Early arrivals result in vehicle queuing and congestion at the destination, while delayed arrivals lead to the idling of Quayside Cranes (QCs) and Yard Cranes (YCs), both of which degrade overall terminal performance. The significance of this system is underscored by studies demonstrating that optimized vehicle dispatching can improve terminal operational efficiency by up to 30% [2], [3]. Consequently, VDS represents a critical leverage point for enhancing the performance of container terminals.

However, the commercialization of VDS has been held back as it is challenging to deploy them across different terminals. This difficulty arises from variations in network layouts, resource and task configuration, and operational requirements, each of which impacts the VDS deployment strategies. Firstly, differing network layouts, which can range from simple unidirectional loops to complex bidirectional grids, mean that a vehicle dispatching logic developed for one topology will be inefficient or invalid in another [4], [5]. Secondly, differences in resource and task configuration, such as fleet size, vehicle types, spatial distribution of tasks, and the placement of cranes [6], require adjustments to the system's core resource allocation settings. Directly deploying VDS without these settings adjustments would lead to degraded performance. Thirdly, differing operational requirements, which govern everything from vehicle priorities to speed limits [7], often require the core logic of the dispatching model to be rewritten for each terminal. Because a VDS is coupled with these terminal-specific characteristics, a system developed for one terminal rarely functions perfectly or, more often, fails entirely in another. Consequently, efforts are required to transfer the VDS across various terminals.

For existing VDSs in the literature, transferring them across terminals is costly due to their heavy reliance on specialists or data. These systems can be broadly divided into

Jia Hu, Junqi Li, and Yuxiong Ji are with the Key Laboratory of Road and Traffic Engineering of the Ministry of Education, Tongji University, Shanghai, China, 201804. (e-mail: hujia@tongji.edu.cn; junqilee@tongji.edu.cn; yxji@tongji.edu.cn)

Weimeng Lin is with COSCO SHIPPING Ports Limited, COSCO Tower No. 658 Dongdaming Road, Shanghai, China (e-mail: lin.weimeng@coscoshipping.com).

Peng Jia is with Collaborative Innovation Center for Transport Studies, Dalian Maritime University, Dalian 116026, Liaoning, China (e-mail: jiapeng@dlmu.edu.cn).

Jintao Lai is with the Department of Control Science and Engineering, Tongji University, No.4800 Cao'an Road, Shanghai, China, 201804. (e-mail: jintao_lai@tongji.edu.cn).

three categories: *i) Experience-driven dispatching systems:* These systems rely on the engineering and operational experience of specialists (engineers) to develop a set of specific dispatching schemes suitable for various operational scenarios (e.g., first-come-first-served, shortest queue first) [8], [9], [10], [11]. Terminal operational scenarios are diverse because they are constantly subject to external variables (e.g., ship arrivals, weather conditions) and internal state changes (e.g., equipment breakdowns, varying traffic congestion). Consequently, this approach often requires multiple engineers, each with specialized experience in different scenarios (e.g., quayside operations, yard management). Transferring such a system requires a team of these highly-paid engineers to manually analyze the new terminal's characteristics and develop a tailored scheme, leading to significant labor costs. *ii) Knowledge-driven dispatching systems:* They rely on the operations research (OR) knowledge of the specialists (scientists) to design a general dispatching approach that covers various operational scenarios [12], [13], [14], [15], [16]. Their transfer requires a scientist-level specialist with a rare and advanced skill set. The scientist should have a deep understanding of OR theory, as well as the specific knowledge of the new terminal's operational characteristics, to correctly develop a vehicle dispatching scheme. This requirement for specialists with such cross-domain knowledge makes them scarce and commands extremely high compensation, thereby resulting in extremely high labor costs for transferability. *iii) Data-driven dispatching systems:* These systems are typically developed using machine learning models, particularly Reinforcement Learning (RL) and its deep learning variants [5], [17], [18], [19], [20]. They demand large volumes of high-quality, terminal-specific data for effective training and validation. If the VDS is transferred to a new terminal, the machine learning model must be retrained on the new terminal's unique operational data to maintain performance. For new terminals, the cost required to collect, cleanse, and label this high-volume data presents a major financial barrier. In summary, transferring the existing VDSs is faced with expertise and data bottlenecks, resulting in high cost.

Another challenge of transferring these VDSs is the time-consuming deployment process required for repetitive manual checks and refinements. The deployment process is typically fragmented to satisfy the requirements of multiple stakeholders. For example, three common stakeholders can be involved in the deployment process: operational staff propose a new operational demand, which is then passed to an optimization specialist for model reformulation, and subsequently to a software engineer for implementation and testing [6], [13], [21]. These handoffs between stakeholders introduce significant communication overhead and potential for misinterpretation, prolonging the deployment process. More importantly, repetitive manual trials and errors are usually required to refine the VDS deployment. They can cost a lot of manual effort and further prolong the deployment process. For instance, humans must manually check the generated code for syntax errors, test the proposed dispatching system, and iteratively refine it. This time-consuming, manual validation loop can span weeks or even months. As a result, this fragmented workflow and iterative refinement process result in long deployment cycles, creating a severe process bottleneck for fast deployment.

Fortunately, the recent emergence of Large Language Models (LLMs) provides a possible opportunity to achieve cheap and fast VDS transferability. LLMs' vast pre-trained knowledge holds the potential to reduce the reliance on human specialists [22], [23], [24], thereby lowering deployment costs. Besides, their powerful contextual understanding and reasoning capabilities support the automated interpretation of dispatching requirements [25], [26], [27], [28], enabling the rapid, automated transferring and configuration of the entire dispatching system for a new environment. These two attributes can be utilized to address the bottlenecks identified above.

However, applying LLMs to the domain of transferring VDSs across ACTs still faces critical challenges that prevent them from being a reliable solution:

*i) Port operation specialists are still required to cope with the unreliable reasoning of LLMs.* The transfer of VDS is a complex, multi-step task requiring long-chain reasoning. A general-purpose LLM, when faced with such an intensive reasoning task, is susceptible to the inherent instability of its long-chain inference. In this case, it often produces logically flawed or incomplete outputs [29], [30]. Consequently, a port operations specialist is still required to guide the LLM's reasoning and validate its answer, ultimately failing to eliminate the human expertise bottleneck.

*ii) Much specific-domain data is required to specialize the LLMs in the domain of vehicle dispatching in ACTs.* Although LLMs have a broad knowledge base, they may lack the high-fidelity, domain-specific knowledge of vehicle dispatching. This gap frequently results in "hallucinations", which means the generation of plausible but inaccurate outputs [31], [32]. To improve the accuracy of generated answers, the LLM must be specialized in the domain of interest. The current mainstream approach for such specialization is to fine-tune a general-purpose LLM to be a specialized one. However, this requires a large volume of domain-specific data [33], [34]. Thus, it is still urgent to address the data bottleneck inherent in the VDS transfer.

*iii) Directly utilizing the LLMs to transfer VDS across terminals does not really save enough deployment time.* This is because LLMs are fundamentally passive text generators and lack an autonomous execution capability to interact with external tools [27], [28], [35], [36]. Therefore, they require a human-in-the-loop to execute the generated code, validate the solution's correctness, and provide corrective feedback. Because LLMs' answers are not always correct on the first attempt, frequent human interventions are often required to validate and correct the answers. This dependency on manual intervention means that directly using an LLM does not eliminate the time-consuming manual deployment process, thereby failing to solve the deployment time bottleneck.

To bridge these gaps, this paper proposes PortAgent, an LLM-driven agent for transferring VDS across various ACTs. To the best of our knowledge, this is the first attempt at designing such an agent for vehicle dispatching in ACTs. The proposed agent exhibits three key features:
- No need for port operations specialists;
- Low need of data;
- Fast deployment.

## II. Highlights

To enable the aforementioned features, the following contributions of this study are highlighted:

**A virtual expert team to avoid the involvement of a port operations specialist.** To overcome the expertise bottleneck, the agent constructs a Virtual Expert Team (VET) to fully harness the reasoning capabilities of LLMs. This team, composed of specialized experts including a Knowledge Retriever, Modeler, Coder, and Debugger, is activated via role-prompting. Each expert is instructed to focus only on its specific sub-task, effectively decomposing the complex problem into a series of shorter reasoning chains. This approach significantly mitigates the risk of logical flaws common in long-chain reasoning, thereby improving VDS solution reliability and demonstrating the capability to replace human specialists.

**Few-shot example learning for low data.** To address the data bottleneck, few-shot learning is employed, enabling the agent to adapt quickly with only a limited number of domain-specific examples. Through this approach, the experts are able to learn VDS-domain knowledge from a few VDS examples. This is facilitated by integrating a Retrieval-Augmented Generation (RAG) mechanism, which draws VDS examples from a curated knowledge base. By leveraging few-shot learning, the agent retrieves and incorporates the most relevant examples into the LLM's context window, allowing for rapid in-context adaptation. As a result, the agent gains grounding in precise domain expertise through few-shot examples, enabling it to generalize effectively to diverse and new terminal scenarios without relying on extensive datasets.

**An automatic VDS design workflow for fast deployment.** To address the process bottleneck, an automatic dispatching workflow is designed. This workflow integrates code generation, execution, and refinement into a closed loop. This is achieved through a self-correction mechanism inspired by the LLM Reflexion framework, which focuses on learning from past failures to improve subsequent attempts. Specifically, this mechanism enables the agent to execute its generated code, reflect on any resulting error feedback to perform a root-cause analysis, and then provide a correction instruction to guide the next generation attempt. This autonomous cycle automates the process of manual checks and refinements, drastically reducing deployment time.

## III. Problem Statement

The problem addressed in this paper is to find an optimal VDS solution for a new terminal environment. This task leads to a VDS searching problem (**P0**): finding an optimal VDS solution from the entire space of possible solutions that maximizes the terminal's operational performance.

**VDS searching problem definition**

Let $\mathbb{E}$ be the set of terminal environments and $\mathbb{P}$ be the set of all possible VDS solutions. They are defined as:

$$\mathbb{E} = \{\mathcal{E}_i | \mathcal{E}_i \text{ is the environment for terminal } i, \quad i \in Z^+\} \quad (1)$$

$$\mathbb{P} = \{p(\mathcal{E}_i) | p(\mathcal{E}_i) \text{ is the VDS solution for terminal } i, \quad i \in Z^+\} \quad (2)$$

Therefore, **P0** is to find an optimal VDS solution, $p^*(\mathcal{E}) \in \mathbb{P}$, for a given environment $\mathcal{E} \in \mathbb{E}$ that maximizes a performance functional $\mathcal{J}$:

$$p^*(\mathcal{E}) = \underset{p \in \mathbb{P}}{\mathrm{argmax}}\, \mathcal{J}(p(\mathcal{E}), \mathcal{E}) \quad (3)$$

where $\mathcal{J}: \mathbb{P} \times \mathbb{E} \to \mathbb{R}$ is the performance functional, a mapping from a given VDS solution and an environment to a real-valued performance metric (e.g., throughput, makespan), $p^*(\mathcal{E})$ is the optimal VDS solution for the given environment $\mathcal{E}$ subject to a set of rules, and $p(\mathcal{E})$ is a feasible VDS solution for environment $\mathcal{E}$:

$$p(\mathcal{E}) = \left\{x^*(\mathcal{E}) = \underset{x \in D}{\mathrm{argmax}}\, Z(x|\mathcal{E}) \,\middle|\, G_k(x|\mathcal{E}) \leq 0,\ \forall k \in Z^+\right\} \quad (4)$$

where $x$ are the decision variables of the VDS (e.g., vehicle route choice variables, task assignment variables, etc.), $D$ is the domain of these decision variables, $Z(x|\mathcal{E})$ is the objective function measuring the VDS's performance under the environment $\mathcal{E}$, $G_k(x|\mathcal{E})$ represents the set of operational constraints derived from the specific terminal environment $\mathcal{E}$, and $x^*(\mathcal{E})$ is the optimal decisions for VDS under the environment $\mathcal{E}$.

**VDS transferring problem definition**

Since directly searching the infinite space $\mathbb{P}$ is intractable, the problem **P0** is unsolvable. Therefore, **P0** is reformulated into a solvable transferring problem (**P1**). Instead of searching the entire space, **P1** aims to generate a high-performance VDS solution by transferring knowledge from existing and finite VDS solutions.

This transfer process is governed by a transfer operator $\mathcal{F}$. $\mathcal{F}$ is a mapping from the environment $\mathbb{E}$ and knowledge base spaces $\mathbb{K}$ to the VDS solution space $\mathbb{P}$:

$$\mathcal{F}: \mathbb{E} \times \mathbb{K} \to \mathbb{P} \quad (5)$$

The transfer operator is then defined as:

$$\mathcal{F}(\mathcal{E}, \mathcal{K}) = p(\mathcal{E}), \mathcal{K} \subseteq \mathbb{K} \quad (6)$$

where $\mathcal{K}$ is the utilized knowledge, a subset retrieved from the overall VDS knowledge base $\mathbb{K}$. $\mathbb{K}$ is a finite set of tuples pairing known environments with their optimal VDS solutions, defined by:

$$\mathbb{K} = \{(\mathcal{E}_i, p^*(\mathcal{E}_i)) | i \in Z^+\} \quad (7)$$

By substituting (6) into (3), **P1** is formally defined as finding the optimal transfer operator $\mathcal{F}$ that maximizes the terminal performance after the transfer process, according to the performance functional $\mathcal{J}$:

$$\mathcal{F}^*(\mathcal{E}, \mathcal{K}) = \underset{\mathcal{F}}{\mathrm{argmax}}\, \mathcal{J}(\mathcal{F}(\mathcal{E}, \mathcal{K}), \mathcal{E}) = p^*(\mathcal{E}) \quad (8)$$

In this work, PortAgent is developed to find the optimal transfer operator $\mathcal{F}^*$ for ensuring VDS performance.

## IV. Methodology

This section details the methodology employed in this study. We first introduce PortAgent's modular architecture, followed by a formal description of the structured inputs and the details of each module.

*A. The PortAgent Architecture*

The architecture of PortAgent is illustrated in Fig. 1. It is built upon three foundational designs that directly address the three transferability bottlenecks:

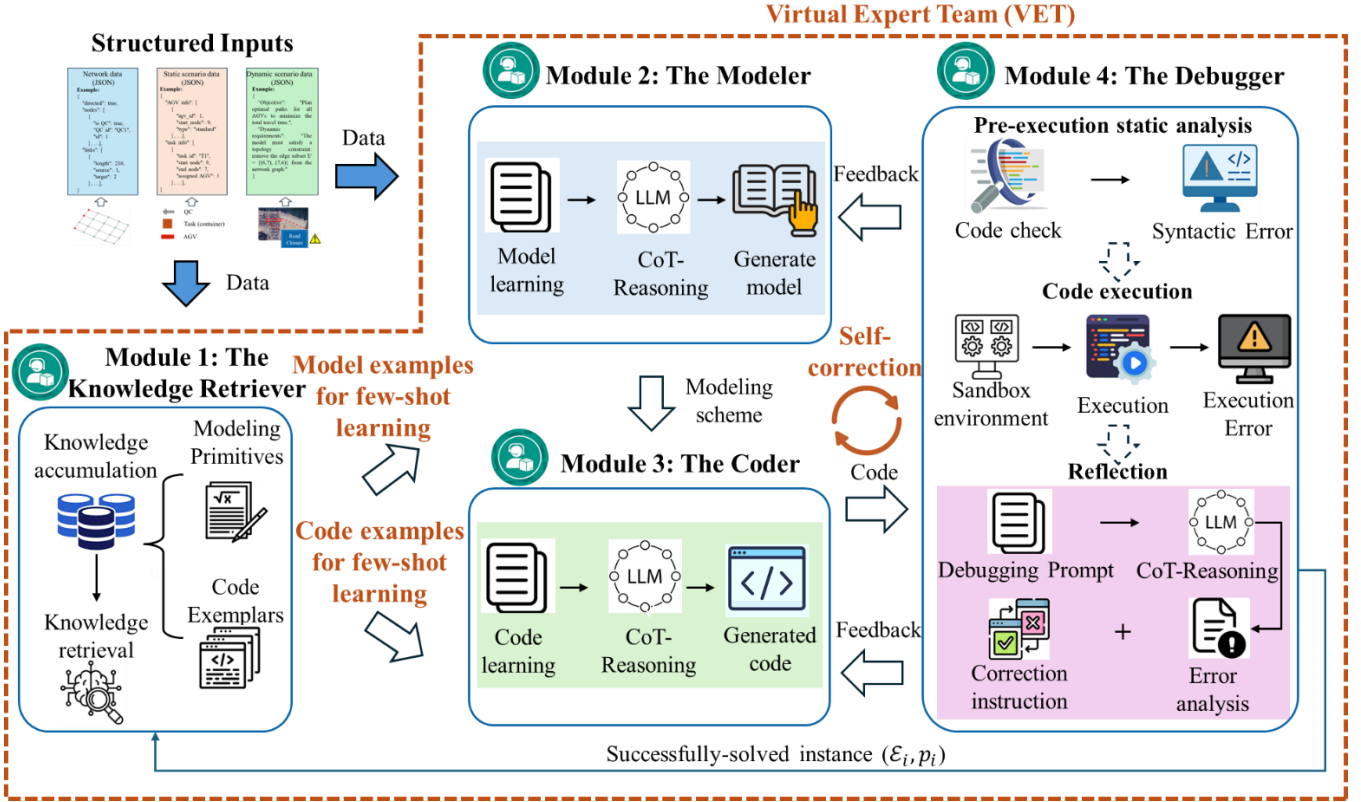

Fig. 1 The architecture of PortAgent.

**VET**: This design constructs a team composed of specialized experts to eliminate reliance on human specialists. The "experts" are not multiple LLMs but rather specialized competencies activated within a single foundational LLM instance through a Role-Prompting Mechanism [37]. Each expert focuses only on a specific sub-task within the VDS transfer workflow.

**Few-shot example learning**: This design leverages the LLM's in-context learning capabilities to achieve data-efficient learning for the experts. It utilizes the RAG mechanism [38] to retrieve the most relevant domain knowledge (e.g. model and code examples) and inject it into the LLM's context.

**Self-correction**: This design enables a closed-loop validation and refinement process for human-free deployment. If an error is detected, the workflow does not terminate. Instead, it generates correction instructions and initiates a new generation attempt.

These designs are implemented through a modular workflow executed by four specialized virtual experts. The sequential functions of these modules are defined as follows:

**Module 1 (The Knowledge Retriever):** This expert operationalizes the knowledge base $\mathbb{K}$. It accumulates the VDS knowledge and executes a targeted RAG retrieval, fetching the most relevant modeling primitives and code exemplars ($\mathcal{K}$). The expert provides the domain knowledge for the subsequent experts.

**Module 2 (The Modeler):** The Modeler learns the retrieved modeling knowledge and conducts a Chain-of-Thought (CoT) reasoning to translate the structured environment inputs $\mathcal{E}$ into a formal natural language modeling scheme.

**Module 3 (The Coder):** The Coder's function is to translate the Modeler's modeling scheme into an executable Python script, yielding the candidate solution $p(\mathcal{E})$. It is achieved by learning code examples and CoT reasoning.

**Module 4 (The Debugger):** The Debugger's role is to validate and correct the candidate solution $p(\mathcal{E})$. It performs static analysis and sandboxed execution to detect any syntactic or runtime errors. Upon detecting an error, it performs reflection to analyze the root cause and generates feedback directed to Module 2 and Module 3, initiating a correction cycle. If the instance is successfully solved, the resulting pair $(\mathcal{E}_i, p^*(\mathcal{E}_i))$ is integrated back into Module 1's knowledge base $\mathbb{K}$, thereby enhancing the agent's future capability.

*B. Structured Inputs*

The input of a terminal environment ($\mathcal{E} \in \mathbb{E}$) is structured across three distinct JSON files, as illustrated in Fig. 2. It is defined as a tuple $\mathcal{E} = (\mathcal{E}_{Net}, \mathcal{E}_{Config}, \mathcal{E}_{Reqs})$:

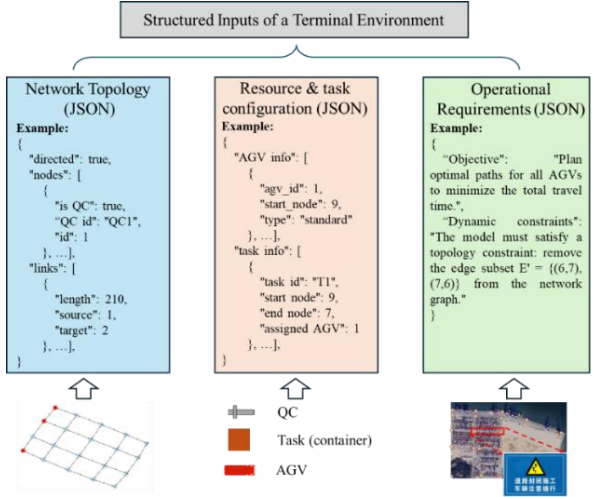

Fig. 2 The structured input of a terminal environment.

**Network Topology ($\mathcal{E}_{Net}$):** It defines the terminal's network topology. It contains a list of nodes with their properties and a list of edges, each specifying its source, target, and length.

**Resource and task configuration ($\mathcal{E}_{Config}$):** This file includes the AGV and task information.

**Operational Requirements ($\mathcal{E}_{Reqs}$):** This file introduces the local rules and dynamic constraints communicated in natural language. These requirements translate into specific constraints $G_k(x|\mathcal{E})$.

### C. Module 1: The Knowledge Retriever

The Knowledge Retriever is the initial expert within the VET workflow. Its responsibility is to manage domain knowledge, executing two key operations: knowledge accumulation and retrieval.

*Knowledge accumulation*

The expert accumulates the VDS knowledge via a knowledge base. This bridges the semantic gap between natural language requirements and formal code. The knowledge base contains two components, as shown in Fig. 3:

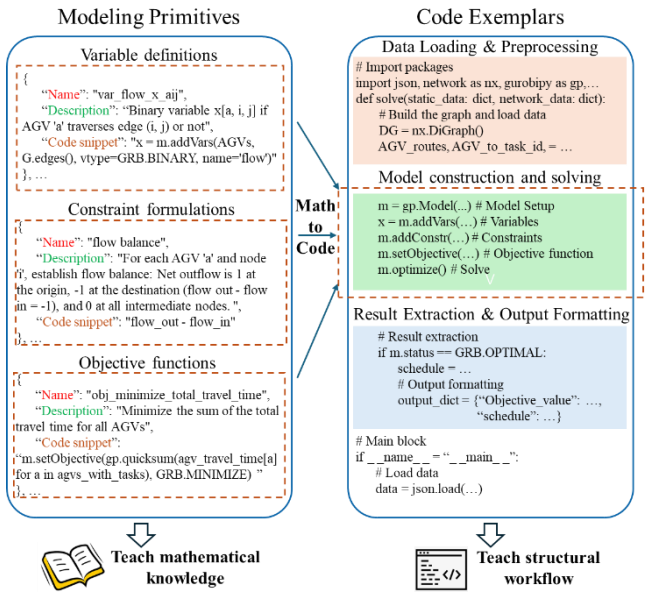

Fig. 3 The structure of the knowledge base.

**Modeling primitives:** This component teaches the agent the basic mathematical primitives of VDS. The primitives are organized as a structured dictionary defining the elements for standard optimization problems, including:

- Variable definitions: Structured definitions for variables.
- Constraint formulations: Canonical formulations for foundational constraints (e.g., flow balance, initial and terminal conditions).
- Objective functions: Standard objectives, such as the minimization of total travel time.

**Code exemplars:** This component consists of specialist-written Python scripts specifically tailored for designing VDSs. Its primary role is not to solve the specific target problem directly, but to teach the agent the overarching structural workflow. It demonstrates best practices for the entire workflow, including data loading and preprocessing, model construction and solving, and structured result extraction and output formatting.

*Knowledge retrieval*

The Knowledge Retriever executes a targeted retrieval process via RAG to ensure the high context fidelity required for effective knowledge injection. For each system transfer attempt, the Retriever analyzes the user's environment input and queries the knowledge base, executing the RAG mechanism to retrieve the most semantically relevant modeling primitives and code exemplars. These retrieved components are then dynamically assembled and injected into the prompt's context window, creating a rich, context-aware prompt that is passed to the Modeler and Coder.

### D. Module 2: The Modeler

Following the knowledge grounding facilitated by the Knowledge Retriever, the Modeler translates the terminal environment inputs and the retrieved knowledge into a mathematical modeling scheme.

As illustrated in Fig. 4, the Modeler receives the RAG-augmented prompt, which contains the structured input data, retrieved modeling primitives, and the correction instruction feedback by the Debugger (Module 4). Upon receiving this, the Modeler engages in two core cognitive processes:

*i) Model learning:* The Modeler utilizes the retrieved model examples to understand the canonical structure of the VDS problem. It articulates how to adapt these examples to the new terminal environment, ensuring that the synthesized model adheres to established optimization best practices.

*ii) CoT reasoning:* The Modeler is instructed to generate a natural language "reasoning" text, which serves as the explicit CoT process [39]. This CoT details the step-by-step mathematical plan before any code is synthesized. The Modeler outlines its understanding of the problem, identifies the necessary mathematical components (decision variables, constraints, and objective function), and articulates how it adapts to a new terminal environment by utilizing the retrieved model examples.

The output of this module is the model formulation scheme accompanied by the CoT reasoning process.

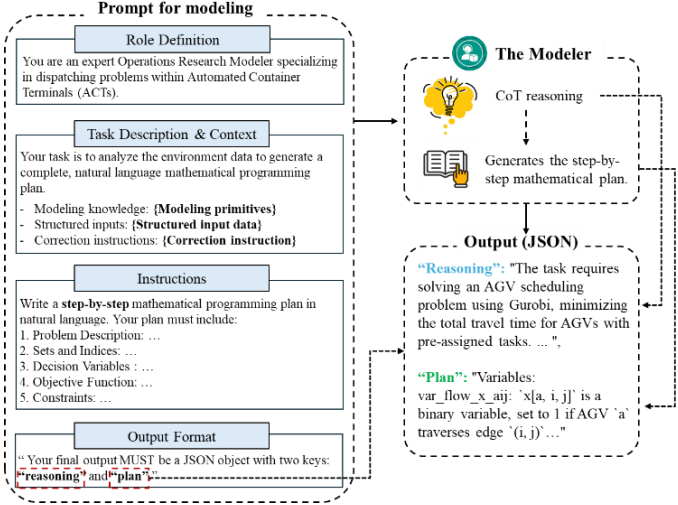

Fig. 4 The process of modeling conducted by The Modeler.

*E. Module 3: The Coder*

Once the mathematical scheme is established, the Coder takes over to synthesize the executable script.

As shown in Fig. 5, the Coder's input context is enriched with the RAG augmented prompt and the natural language model formulation generated by the Modeler. The Coder's task is not to generate code from scratch, but rather to perform a highly accurate assembly process. It utilizes the Modeler's output as a strict logical specification to guide the generation of logical components. The Coder performs two specialized operations:

*i) Code learning*: The Coder relies on the provided code exemplars for the overall procedural structure. This ensures correct implementation of data loading, model initialization, solver execution, and structured result formatting.

*ii) CoT reasoning*: To ensure verifiability and traceability, the Coder generates a CoT reasoning text that details the process of translating the Modeler's abstract modeling scheme into functional code.

The output of this module is the complete, executable Python code script with CoT reasoning process.

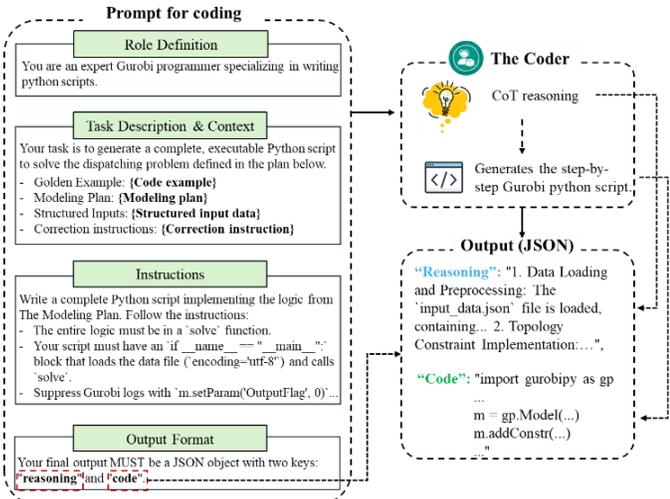

Fig. 5 The process of code generation conducted by The Coder.

*F. Module 4: The Debugger*

The Debugger is responsible for executing, detecting, and resolving errors in the generated code.

The complete process of this module is illustrated in Fig. 6. As shown in the figure, after an initial code generation attempt, the agent enters a systematic loop that includes static code analysis, execution, and, upon failure, reflection. This process continues until a valid, executable solution is produced, or a predefined maximum number of iterations is reached.

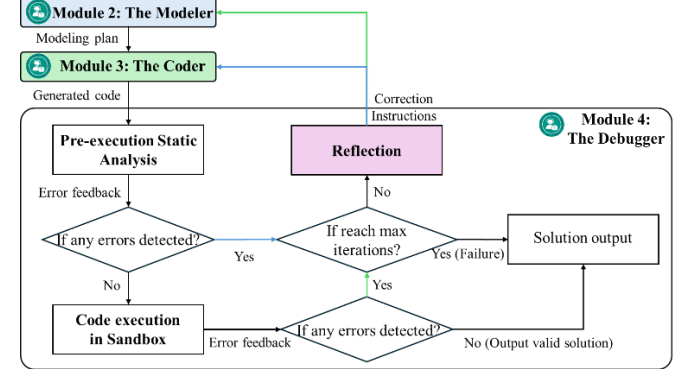

Fig. 6 The flowchart of The Debugger.

*Pre-execution static analysis*

Pre-execution static analysis is the first attempt at code validation. Before running the generated script, the agent performs a lightweight check using Abstract Syntax Trees (AST) [40]. This technique parses the code into a structured tree representation, allowing the Debugger to verify its structural integrity and confirm the presence of required modeling components. This step can pre-emptively catch certain structural or syntactic flaws, thereby reducing the frequency of costly execution attempts.

*Execution and error feedback*

If the pre-execution static analysis passes, the agent executes the generated Python script in a sandboxed environment. The system is designed to capture all forms of feedback from this execution attempt, which can be categorized into two error types: i) Syntax errors: Failures where the Python interpreter cannot parse the code. ii) Runtime errors: Errors that occur during code execution, including exceptions raised by the solver (e.g., invalid variable indexing or incorrect handling of input data).

The error message and the failed code are then fed into the reflection phase.

*Reflection*

Upon detecting any error, the reflection phase is initiated, which is fundamentally inspired by the LLM Reflexion framework that leverages past failures to improve subsequent attempts [41]. As shown in Fig. 7, the Debugger's task is not to fix the code directly but to perform a two-step root-cause analysis that drives the refinement cycle:

*i) Diagnosis.* The expert engages in a focused CoT-based reasoning process. It analyzes the error message in the context of the failed code, hypothesizes the root cause, and proposes a conceptual fix. This ensures a deep understanding of the problem before a solution is prescribed.

ii) Correction. The expert outputs a concise natural-language correction instruction. This correction instruction acts as a direct guide for the Modeler and Coder, instructing them on how to avoid the previous mistake.

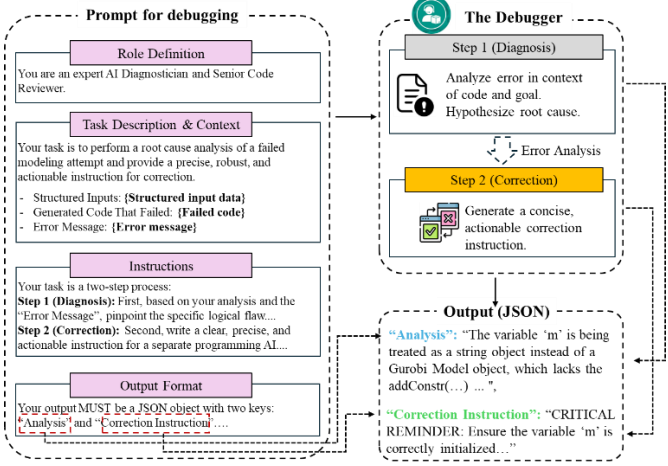

Fig. 7 The process of reflection conducted by The Debugger.

## V. EVALUATION

Experiments have been designed to evaluate the proposed agent. The performance of the agent is evaluated from the following aspects: i) transferability; ii) specialist-free deployment capability; iii) low data requirement; iv) deployment speed.

### A. Experiment Design

To assess the capabilities of the proposed agent, a comprehensive experiment is designed.

*VDS type of interest*

The PortAgent's transferability is evaluated using the Multi-AGV Path Planning (MAPP) problem. MAPP is an ideal VDS type because it represents the critical execution layer in ACTs where dynamic constraints most frequently manifest.

MAPP is defined on a directed graph $G = (N, E)$, representing the terminal's road network. $N$ is the set of nodes (locations) and $E$ is the set of edges (unidirectional road segments), with each edge $(i, j) \in E$ having an associated length $d_{ij}$. Consider a fleet of AGVs $V$, and a set of transportation tasks $T$. Each task is pre-assigned to a specific AGV, defining an origin-destination pair $(s_v, t_v)$ for each vehicle $v \in V$. The objective is to find a set of paths $\{P_v | v \in V\}$ that minimizes the total travel time for the entire fleet, formulated as:

$$\min Z = \sum_{v \in V} \sum_{(i,j) \in E} d_{ij} x_{vij} \quad (9)$$

subject to the following standard flow-balance constraints:

$$\sum_{j \in N | (i,j) \in E} x_{vij} - \sum_{j \in N | (j,i) \in E} x_{vji} = b_{vi}, \forall v \in V, \forall i \in N \quad (10)$$

where $b_{vi}$ is defined as:

$$b_{vi} = \begin{cases} 1, & \text{if } i = s_v \\ -1, & \text{if } i = t_v \\ 0, & \text{otherwise} \end{cases}, \forall v \in V, \forall i \in N \quad (11)$$

The decision variables $x_{vij}$ is a binary variable that equals 1 if AGV $v$ traverses edge $(i, j)$, and 0 otherwise.

*Scenarios for test*

To evaluate the agent's performance against different scenarios for transferring VDSs, three typical scenarios of MAPP are selected for test, as listed in Table 1.

Table 1 Typical scenarios for test.

| Scenario | Description |
|---|---|
| Road closure | An unexpected operational disruption renders a bidirectional road segment impassable (e.g., due to an accident or maintenance). |
| Forbidden roads for specific trucks. | A truck with specific physical attributes (e.g., over-height, excess weight, or excess width) is prohibited from traversing a designated road segment due to infrastructure limitations (e.g., a low bridge) or safety protocols. |
| Designated routes for dangerous goods | A specific task (e.g., T3, carrying dangerous goods) must follow a mandatory and safe sub-path. |

The experimental testbed is built upon a representative port network (30 AGVs, 20 nodes), as shown in Fig. 8. To ensure statistical robustness, for each scenario of the three scenarios, five unique instances were generated using distinct random seeds to sample the AGV origin and destination nodes. This resulted in a total of 15 base scenarios.

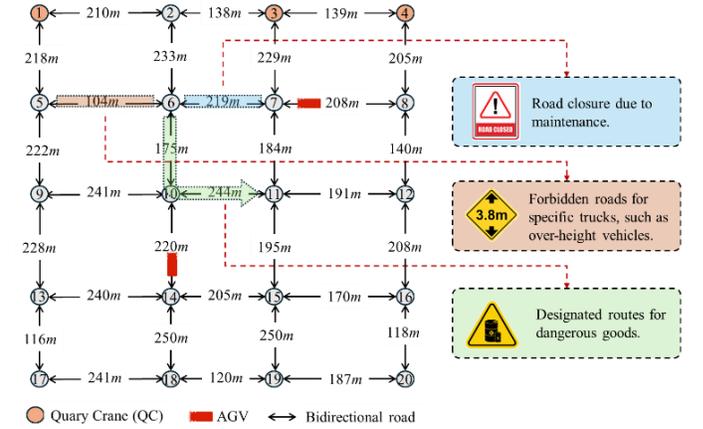

Fig. 8 The network for test.

*Expertise levels of human input*

To investigate the agent's robustness to the expertise levels of human inputs, each tested scenario was articulated using three distinct linguistic styles, simulating different levels of operational expertise. Table 2 summarizes the three styles of expertise level.

Table 2 Three styles of expertise levels.

| Expertise level | Input style | Example phrasing (e.g., road closure) | Purpose |
|---|---|---|---|
| Technician-level | Informal, physical description. Uses everyday language, describing the physical situation rather than formal rules. | "The road between Node 6 and Node 7 is closed." | Simulates a user with no formal OR training. |
| Engineer-level | Clear, operational terminology. Uses unambiguous operational terms but avoids mathematical formalism. | "The road segment connecting (6, 7) is closed in both directions." | Simulates an expert user with good operational knowledge. |
| Scientist-level | Formal, mathematical specifications. Uses precise terminology from operations research and mathematics. | "A global network constraint E′=E\{(6,7), (7,6)}." | Simulates a scientist with OR modeling knowledge. |

The three distinct linguistic styles resulted in a comprehensive testbed of 15 base scenarios × 3 expertise levels = 45 unique test instances. The complete prompts for each scenario and expertise level are provided in Appendix A.

*Benchmark method*

The proposed agent was benchmarked against solutions derived from a traditional specialist-driven method. For each of the test instances, a human expert with a background in OR manually translated the problem description into a precise mathematical formulation and implemented it in a Python script using the Gurobi solver. This manually-coded implementation serves as the ground truth, representing the optimal solution for each test instance.

*Measures of Effectiveness (MoEs)*

The performance of the agent is quantified from two aspects: solution correctness and computational efficiency.

**The MoEs of solution correctness:**

i) Code executability rate (CER): This metric assesses the agent's fundamental capability to generate syntactically valid and executable Python code. A run is deemed executable if it completes without any runtime errors. The CER is defined as:

$$CER = \frac{N_{executed}}{N_{total}} \quad (12)$$

where $N_{executed}$ is the number of test instances that generate successfully executed scripts and $N_{total}$ is the total number of test instances.

ii) Solver success rate (SSR): This metric measures the semantic and logical correctness of the formulated optimization model. A case is deemed successful if it matches the objective value of the Gurobi ground truth solution within a predefined tolerance. The SSR is defined as:

$$SSR = \frac{N_{solved}}{N_{total}} \quad (13)$$

where $N_{solved}$ is the number of test instances where the generated script yields a correct solution.

**The MoEs of computational efficiency:**

i) Iterations: This metric quantifies the performance of the agent's self-correction mechanism. It records the number of generation-and-review cycles required to produce a final, error-free, and validated script.

ii) Computation time: This metric measures the end-to-end time required for the agent to proceed from receiving the initial natural language prompt to outputting the final solution, thereby quantifying the overall efficiency of the automated workflow.

*Experimental settings*

All experiments were conducted on a personal computer equipped with a 13th Gen Intel(R) Core (TM) i5-13500H processor and 16 GB of RAM. The agent and the benchmark method were implemented in Python 3.9.

**Foundational LLM**:
- Reasoning engine for the agent: Gemini 2.5 Flash.
- Temperature: 0.
- Communication with the LLM: standard API.

**Agent configuration:**
- The maximum iterations: 3.
- Number of examples for learning: 1.
- Permitted optimality error: 1e-4.

**Optimization tool:**
- Modeling language: Pyomo.
- Solver: Gurobi Optimizer version 12.0.3.
- Solving time limit: 300 *seconds*.

*B. Results*

Results confirm the agent could meet the outlined objectives: i) The agent demonstrates robust transferability, achieving a 100% CER and an SSR ranging from 86.67% to 100% across different dispatching scenarios. ii) It effectively reduces the need for optimization specialists, with even "Technician-level" users achieving an 86.67% average SSR, and shows no statistically significant performance difference across expertise levels. iii) The agent operates with high data efficiency, requiring only a single example to achieve an average SSR of 93.33%. iv) It enables fast deployment, with an average end-to-end computation time of just 83.23 *seconds*.

*Transferability*

**i) Transfer success rate**

Fig. 9 presents the results of the transfer success rate, demonstrating the agent's capability of transferring VDSs across different terminal scenarios. Fig. 9(a) shows the transfer success rate for code execution. It demonstrates that the agent consistently generated syntactically valid and executable code, achieving a perfect 100% CER. Fig. 9(b) shows the transfer success rate for solution correctness, which represents the agent's performance in generating logically correct solutions. The SSR ranged from 86.67% to 100% across the three scenarios, indicating that the agent can reliably transfer its knowledge to solve a variety of unseen terminal scenarios.

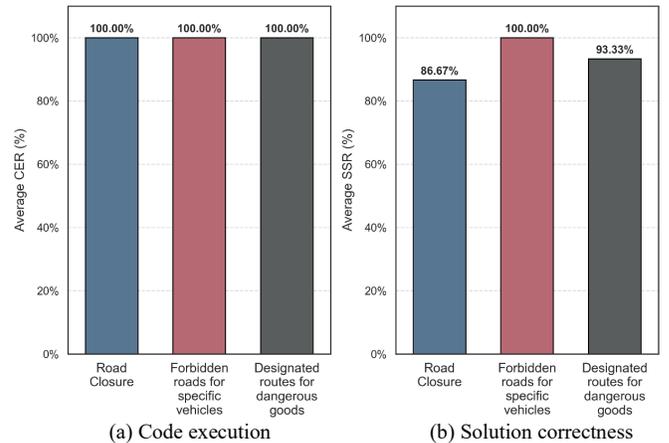

(a) Code execution (b) Solution correctness
Fig. 9 Transfer success rate across different terminal scenarios.

**ii) Root-cause analysis for transfer failures**

To better understand the agent's limitations, a root-cause analysis is performed. Out of the 45 total test instances, the agent successfully solved 42, yielding an overall SSR of 93.33%. The failures of the 3 unsuccessful test instances were categorized as shown in Table 3. As summarized in the table, all failures were uniformly categorized as "Misinterpretation", wherein the generated code executed successfully, yet the resulting solution was incorrect.

Table 3 Error types of the 3 unsolved test instances.

| Error Category | Specific Description | Count |
|---|---|---|
| Misinterpretation | The model did not correctly enforce the bidirectional nature of the road closure constraint. | 2 |
| | The model incorrectly constrained the entire path to the designated route, instead of treating it as a sub-path. | 1 |

This analysis reveals a clear boundary in the agent's capabilities: it cannot autonomously address failures caused by semantic misinterpretation, where the generated code is syntactically perfect but yields a logically flawed solution. The underlying reasons for these failures stem from two factors: i) Ambiguity of user input: The requirements are often communicated by users in natural language. This input can be inherently ambiguous, leading to potential misinterpretation of the user's intended mathematical formulation by the agent. ii) Probabilistic LLM output: As a probabilistic generative model, the underlying LLM exhibits inherent randomness. Even identical inputs may occasionally lead to logical variance in the output (i.e., some random seeds correctly interpret the constraint, while others do not).

Therefore, future research should focus on two corresponding directions: improving semantic understanding to reduce the inherent ambiguity of user inputs and enhancing output consistency to mitigate the effects of probabilistic randomness.

*Specialist-free deployment capability*

This section evaluates the agent's performance when provided with instructions from users with varying expertise levels. The agent's performance was assessed in two aspects: solution correctness and computational efficiency.

**i) Solution correctness with/without specialists**

Fig. 10 presents the agent's solution correctness, as measured by CER and SSR, demonstrating that it achieves robust and consistently high performance irrespective of the user's expertise level. The agent achieved a perfect 100% CER. While minor variations were observed in the SSR, the agent achieved consistently high performance across all expertise levels. Notably, the "Scientist-level" input yielded a perfect 100% SSR. This success is attributable to the unambiguous, formal, and often mathematical phrasing, which minimizes the potential for semantic misinterpretation. Conversely, the SSR for "Technician" and "Engineer" level inputs, while still strong, fell slightly below 100%. This outcome aligns perfectly with the root-cause analysis in the aforementioned: their use of more vague natural language introduces ambiguity, increasing the risk of the LLM misinterpreting the required constraints.

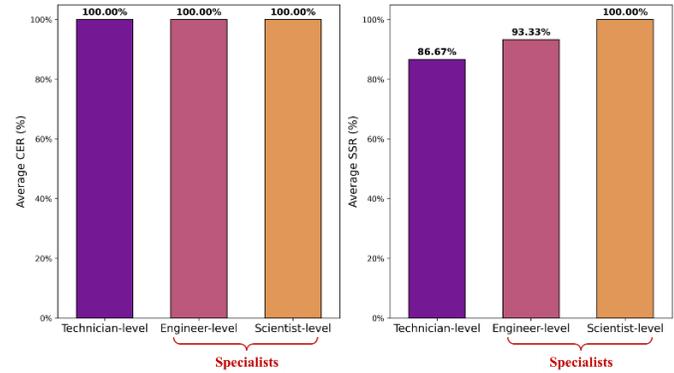

Fig. 10 Solution correctness across different expertise levels.

**ii) Computational efficiency with/without specialists**

Fig. 11 demonstrates that the agent's computational efficiency, quantified by the number of iterations and total computation time, is not significantly impacted by the users' expertise levels. It is observed that "Engineer-level" inputs resulted in slightly more iterations on average. The reason behind this phenomenon is that they are more complex than a "Technician-level" description, introducing subtle syntactical or logical complexities, yet lack the formal precision of a "Scientist-level" input, which makes them more likely to generate initial code with errors that trigger the self-correction mechanism.

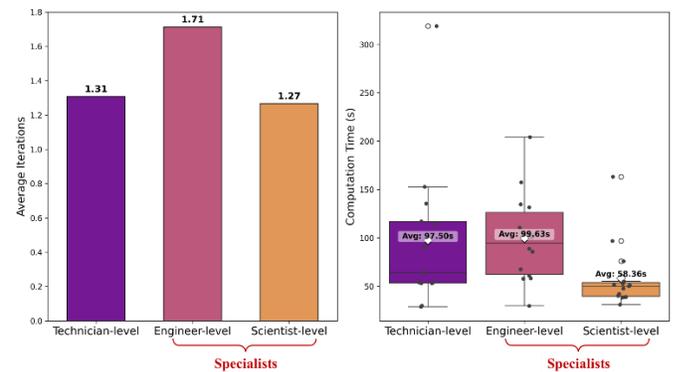

Fig. 11 Computational efficiency across different expertise levels.

**iii) Significance of specialists**

Table 4 provides the results of statistical tests, showing no statistically significant differences in any of the four MoEs across the three expertise levels. This leads to a key finding of this study: the expertise level of the user has no significant impact on the agent's ability to correctly solve the problem, nor on the time required to do so. This quantitatively proves that the proposed agent eliminates the dependency on port operations specialists, including engineers and scientists, thus successfully addressing the expertise bottleneck.

Table 4 Statistical analysis of the impact of expertise level.

| Aspect | MoEs | Test | Statistics | p-value | Significance* |
|---|---|---|---|---|---|
| Solution correctness | CER | Chi-squared ($\chi^2$) | $\chi^2$= 0.0000 | 1.0000 | Not Significant |
| | SSR | | $\chi^2$= 2.1429 | 0.3425 | Not Significant |
| Efficiency | Iterations | ANOVA (F-statistic) | F= 2.3117 | 0.1125 | Not Significant |
| | Computation | | F= 2.6326 | 0.0846 | Not Significant |

| | Time | | | |

*Significance is determined at the p < 0.05 level.*

### Low data requirement

This section evaluates the agent's data efficiency by analyzing its performance with a varying number and types of examples. The expertise level is set as "Engineer-level".

**i) Example quantity needed**

To validate the agent's low data requirement, its performance was evaluated under three conditions: 0-shot, 1-shot, and 3-shot. The 0-shot setting provided no code example, while the 1-shot setting provided a single example solving the classic MAPP model presented in Equations (9) - (11). The 3-shot setting provided three distinct examples, each solving one of the tested dispatching scenarios.

As illustrated in Fig. 12, the results confirm that the agent operates with high data efficiency, as its performance across key metrics saturates with only a single example (1-shot). The 1-shot configuration consistently achieves the highest CER and SSR and the lowest computation time, establishing it as the most effective approach.

A counterintuitive finding emerged when comparing 1-shot to 3-shot learning: providing more examples led to a degradation in performance. This occurs because multiple, slightly different examples can introduce conflicting patterns or "contextual noise," which may confuse the LLM during code generation rather than providing clearer guidance.

Notably, another paradox was observed: while the 1-shot configuration required the most self-correction iterations on average, it was also the fastest in terms of computation time. This suggests that the single, simple example leads to a straightforward initial code generation attempt. When this attempt fails, the resulting errors are typically simple, allowing the self-correction mechanism to diagnose and resolve them rapidly. This highlights that the 1-shot approach provides the optimal balance between sufficient guidance and minimal complexity, enabling both high accuracy and high efficiency.

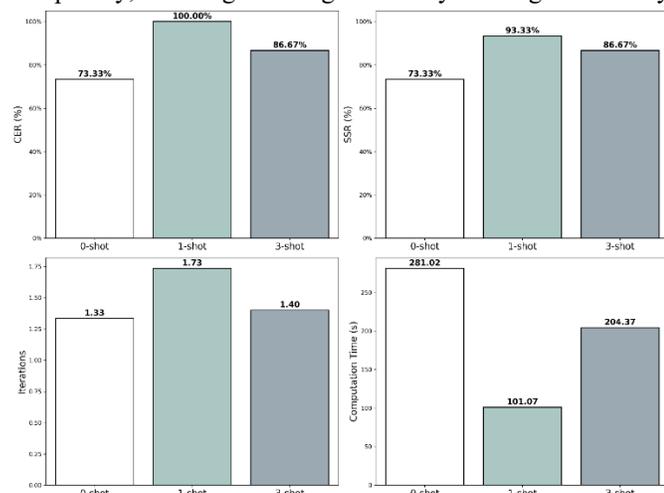

Fig. 12 Impact of number of examples across all MoEs.

**ii) Example type required**

The requirement of the single example type (1-shot configuration) is investigated. An experiment was conducted comparing the agent's performance when guided by four different example types: a classic dispatching example versus three customized examples, each pre-solved for one of the three specific testing scenarios.

The results, presented in Fig. 13, are striking. The agent guided by the classic dispatching example achieved superior or joint-best performance across both CER (100%) and SSR (93.33%), significantly outperforming agents guided by examples that were supposedly more relevant to the target scenario. Besides, while the classic example required the most iterations on average (1.73), it was by far the most efficient in computation time (101.07 seconds).

This occurs because the classic example provides a clean structural workflow. It teaches the agent the fundamental workflow without introducing scenario-specific logical knowledge. The agent's initial code is therefore simple, and any necessary corrections are easily localized and computationally fast to resolve. In contrast, a specialized example introduces "contextual noise," causing the LLM to generate more complicated initial code that may be overfit to irrelevant details. When this more complex code fails, the debugging process is more involved, leading to fewer but significantly longer iterations (e.g., the agent guided by the road closure example averaged 300.60 seconds).

This leads to a powerful design principle for LLM-driven code generation agents: the few-shot example should primarily teach the fundamental workflow, not the specific problem logic. This is a more effective and robust strategy for achieving high transferability performance.

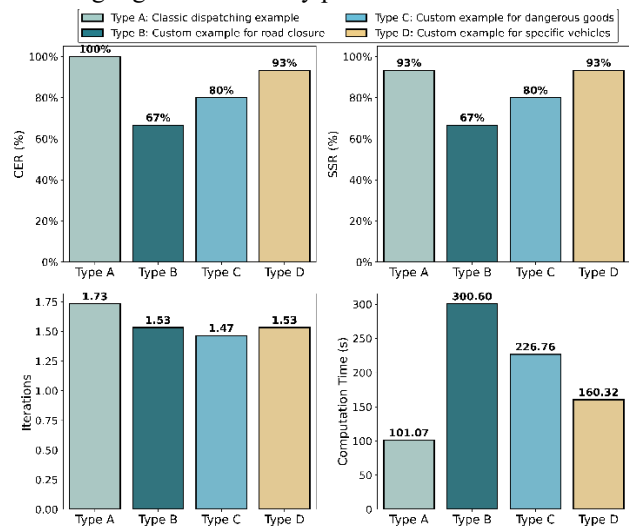

Fig. 13 Performance comparison of the agent when guided by a single basic vs. a specialized example.

### Deployment speed

Table 5 provides a comparison between the agent's automated workflow time and the typical time required for manual specialists, confirming the agent's capability of fast deployment. By automating the end-to-end process of model formulation, coding, and debugging, the agent reduces a task that takes hours or even days to an average of just 83.20 seconds. The ability to generate a validated, executable optimization model in under two minutes validates the agent's critical role in enabling fast deployment and addressing dynamic operational challenges in real-time port environments.

Table 5 Comparison of deployment time.

| Method | Computation time |
|---|---|
| Benchmark (Traditional specialist-driven method) | Several hours to several days. |
| PortAgent | 83.20 seconds (on average) |

*Ablation study*

To isolate and quantify the contributions of the agent's key architectural components, an ablation study is conducted. The performance of the full PortAgent was benchmarked against two degraded configurations, using "Engineer-level" prompts. The ablated configurations were:

i) PortAgent w/o RAG: This configuration removes access to the RAG mechanism, meaning the agent operates without its curated knowledge base, including the modeling primitives and the few-shot code exemplar. This validates the necessity of the knowledge grounding module.

ii) PortAgent w/o self-correction: This configuration restricts the agent to a single, non-iterative generation attempt, effectively disabling the closed-loop reflection and refinement mechanism. This demonstrates the critical role of autonomous debugging.

Table 6 provides the results of the ablation study, demonstrating the significance of both modules. Without the RAG, the agent's performance degrades, with the SSR dropping to a mere 26.7%. This underscores the necessity of the knowledge base for grounding the LLM's reasoning and providing the essential modeling primitives required to formulate valid models. Similarly, without the self-correction, the SSR falls to 33.33%. This demonstrates that the reflection and iterative refinement loop is indispensable. Even when provided with the RAG, the LLM's initial generation attempt often contains errors; the self-correction module is essential for catching and fixing these errors.

Collectively, these results validate the proposed architecture, proving that both RAG (specific-domain knowledge grounding) and the self-correction mechanism (autonomous debugging) are essential and non-redundant components required to achieve the agent's high performance.

Table 6 Ablation experiments on RAG and self-correction mechanisms.

| Method | CER | SSR |
|---|---|---|
| PortAgent with full modules | **100%** | **93.3%** |
| PortAgent w/o RAG | 40.0% | 26.7% |
| PortAgent w/o self-correction | 33.33% | 33.33% |

## VI. CONCLUSION AND FUTURE WORK

This paper proposes an LLM-driven vehicle dispatching agent, PortAgent. This agent automates the transfer of VDS across various ACTs. It has the following features: (i) no need for a port operations specialist; (ii) low need of data; (iii) fast deployment. To validate these features, a comprehensive evaluation was conducted across various vehicle dispatching scenarios. PortAgent has been compared against the traditional specialist-driven method. The results show that:

i) The proposed agent demonstrates robust transferability, achieving a 100% CER and an SSR ranging from 86.67% to 100% across different dispatching scenarios.

ii) It effectively reduces the need for optimization specialists. It shows no statistically significant difference in performance across user expertise levels, with even "Technician-level" users achieving an 86.67% average SSR.

iii) The agent operates with high data efficiency, requiring only one example to achieve an average SSR of 93.33%. It is suggested that.

iv) It enables fast deployment, reducing a process that typically takes hours or days to an average end-to-end computation time of just 83.23 seconds.

v) The results also revealed two interesting methodological insights crucial for LLM-driven agent design:

- *Less is more:* Increasing the number of examples for few-shot learning can be detrimental to solution correctness and computational efficiency, indicating that optimal performance relies on example quality over quantity.
- *Fundamental is the key:* Few-shot example learning is most effective when targeted towards learning the fundamental knowledge, rather than attempting to directly find complex and specific problem answers. This approach can maximize the reliability of the transferred VDSs.

**Future work:** While PortAgent marks a significant advance, one limitation remains: semantic misinterpretation. The limitation stems from the inherent ambiguity of natural language user inputs and the probabilistic randomness of LLM outputs. Accordingly, our future research will develop novel methods to enhance the agent's semantic understanding and ensure greater output consistency.

## APPENDIX

*A. Prompts for Test Scenarios*

Table 7 provides the full text of the natural language prompts used for each test scenario across the three defined expertise levels.

Table 7 The full description of vehicle dispatching scenarios.

| Scenario | Expertise level | Natural language prompts |
|---|---|---|
| Road Closure | Technician-level | "That road between node 6 and node 7 can't be used today." |
| | Engineer-level | "Attention: The bidirectional road segment connecting nodes (6, 7) is completely closed." |
| | Scientist-level | "The model must satisfy a topology constraint: remove the edge subset E' = {(6,7), (7,6)} from the network graph." |
| Forbidden roads for specific trucks | Technician-level | "AGV-4 in the fleet is one of those extra-tall ones; it can't get under the low bridge between node 5 and node 6." |
| | Engineer-level | "Attention: AGV-4 in the fleet is an over-height vehicle and cannot pass through the bidirectional height-restricted gantry connecting (5, 6)." |
| | Scientist-level | "A vehicle-path compatibility constraint must be enforced: for v=4, the decision variable $x_{ve}$ must be 0 for all $e$ in {(5,6), (6,5)}." |

| Designated routes for dangerous goods | Technician-level | "The container for T3 has dangerous goods, so it has to stick to the safe route: go from 6 to 10, then from 10 to 11. No exceptions." |
| --- | --- | --- |
| | Engineer-level | "Task T3 involves dangerous goods and must follow the designated one-way safety corridor (6->10->11)." |
| | Scientist-level | "A mandatory subpath constraint must be applied to the AGV assigned to task T3, ensuring its solution path contains the subsequence (6, 10, 11)." |

## DATA AVAILABILITY

The data and code supporting the findings of this study are available from the corresponding author upon reasonable request.